\documentclass[a4paper,twoside]{article}

\usepackage{epsfig}
\usepackage{subcaption}
\usepackage{calc}
\usepackage{amssymb}
\usepackage{amstext}
\usepackage{amsmath}
\usepackage{amsthm}
\usepackage{multicol}
\usepackage{pslatex}
\usepackage{apalike}
\usepackage[ruled]{algorithm2e}
\usepackage{url}
\usepackage[cdot,amssymb]{SIunits}

\usepackage{SCITEPRESS}  
\begin{document}
	
	\title{A Hybrid Approach for Reinforcement Learning Using Virtual Policy Gradient for Balancing an Inverted Pendulum}
	\author{\authorname{Dylan Bates\sup{1}\includegraphics[height=0.3cm]{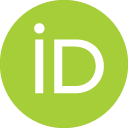}\thanks{\includegraphics[height=0.3cm]{orcid.png}~https://orcid.org/0000-0002-2706-443X}}
		\affiliation{\sup{1}Center for Research in Scientific Computation, North Carolina State University, Raleigh, NC, USA}
		\email{dwbates@ncsu.edu}
	}
	
	\onecolumn \maketitle \normalsize \setcounter{footnote}{0} \vfill
	
	\section{\uppercase{RESEARCH PROBLEM}}
\begin{figure}[bt]
		
		\begin{center}
			\mbox{
				\includegraphics[width=\columnwidth]{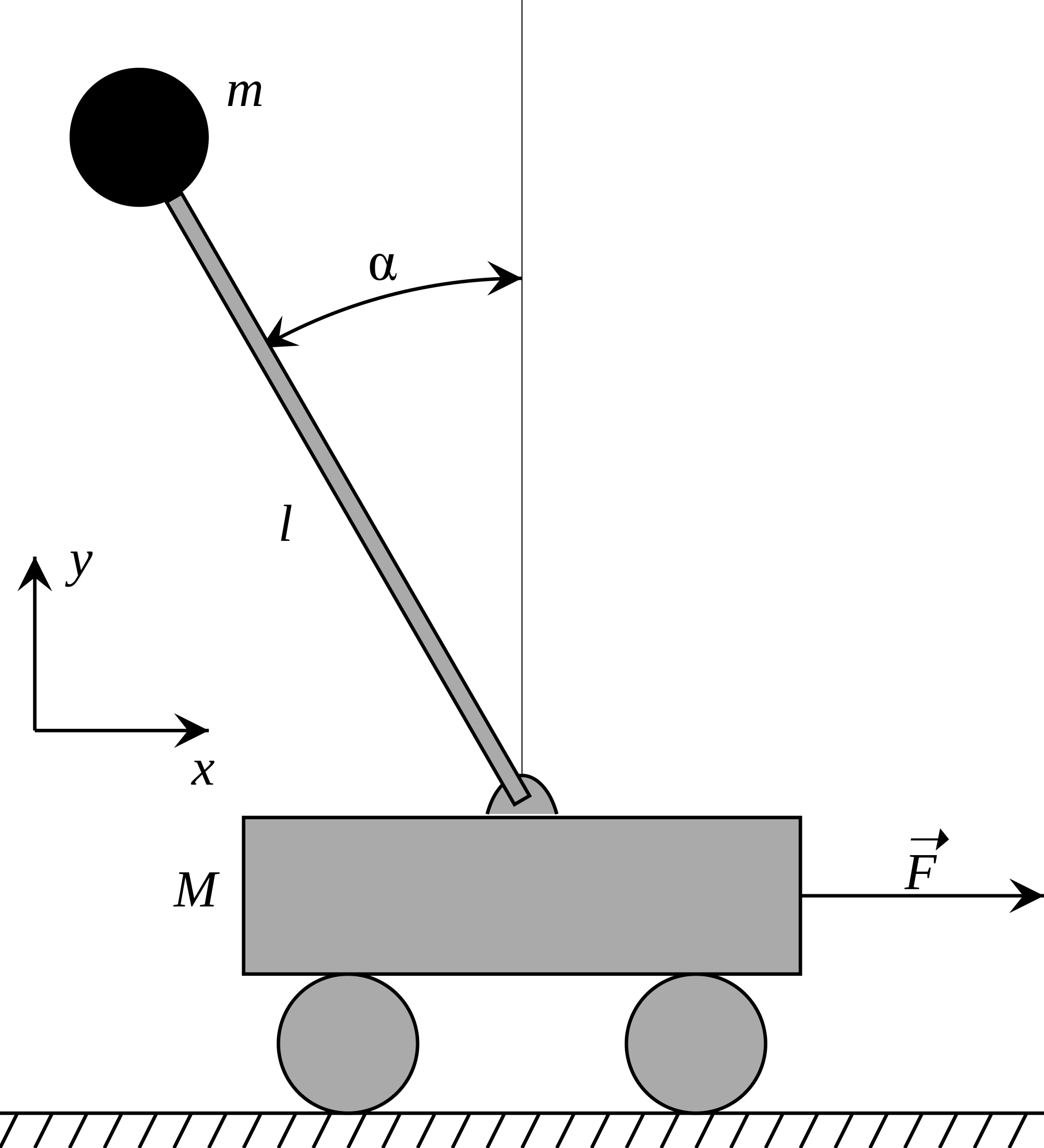}
			}
		\end{center}
		
		\caption{SIP Model. The cart is only free to move with 1 degree of freedom.}
		\label{SIP}	
	\end{figure}

	Reinforcement learning is the process of using trial-and-error with finite rewards to encourage an agent to take the optimal action when presented with the state of its environment. Our agent is a single-inverted pendulum, attached to a cart on a one-dimensional track, with a 4-dimensional continuous state space: $[x,\alpha,\dot{x},\dot{\alpha}]$ -- corresponding to the $x$-position (right positive), angle (counterclockwise positive from 0 vertical), $x$-velocity, and angular velocity, as indicated in Figure~\ref{SIP}. 
	
	The inverted pendulum is an underactuated system with an unstable
	equilibrium. By supplying voltage to the motor, which can only move the cart
	left or right along a finite length of track, the goal is to balance the pendulum
	directly above the cart. Ideally, the system will also be robust to sensor noise,
	motor play, and forced disturbances (i.e. being gently tapped).

	The action space is the continuous range of voltages available to the motor, ranging from $-10$V to $+10$V, moving the cart left or right, respectively. The cart will also move, according to the physics of the system, even if no voltage is applied. Our goal is to apply the appropriate voltage to the motor in order to balance the pendulum vertically above the cart, without running the cart off the track. The novel approach proposed here is to train the model entirely through simulation, virtually in Python. The trained model is then transferred to the real world agent, where the inverted pendulum is balanced over the physical cart.
	

	\section{\uppercase{OUTLINE OF OBJECTIVES}}
	\label{sec:objectives}
	The primary objective of this project is to use a variety of reinforcement learning algorithms to train a realistic simulation of a virtual pole to balance itself. The trained function or neural network will then be used to balance a real pole in real time. Ultimately, training through simulation can speed up training time and increase robustness.
	
	Moving forward, the objective is to test a variety of control algorithms for
	training speed and robustness, in order to find a robust algorithm that can learn to consistently balance on its own. A variety of data-efficient algorithms will be benchmarked, and eventually implemented in other applications like a double inverted pendulum.
	
	Ultimately, the idea of using reinforcement learning to train a single inverted pendulum is a proof of concept that could make other reinforcement learning applications more efficient, letting computers quickly and efficiently train models before implementing them in the real world. Eventually, such an algorithm could be used in a variety of simulatable real-world applications like robotics, self-driving cars, and reusable rockets.

	\section{\uppercase{STATE OF THE ART}}
	\subsection{Literature Review}\label{lit}
	Single inverted pendulums are ubiquitous benchmark problems used for testing various control schemes. Because the model can be simulated to an extremely high degree of accuracy, computer simulations are often used to test these controls instead of real-world pendulums, which can be expensive to obtain, time-consuming to set up, and labourious to maintain. Due to their ubiquity, there have been a plethora of classic control techniques used to balance them. These techniques generally fall into three categories.
	
	\subsubsection{Reinforcement Learning on Virtual Pole}
	Using a simplified environment, a handful of reinforcement learning techniques have been used to balance OpenAI's Cartpole environment\footnote{\url{https://gym.openai.com/envs/CartPole-v0/}}. The original environment defined in \cite{Cartpole} used an Adaptive Critic method to greatly improve on the Boxes algorithm used in \cite{Boxes} to teach a computer-analogue to play noughts and crosses (tic-tac-toe). The discretization of the state space into 225 (refined in 1983 to 162) regions or ``boxes'' was intended to create several hundred independent sub-problems that could each be solved, finding the optimal action to take any time the pole was in a specific box. The algorithm is remarkably simple: once the pole state enters, a \textit{local demon} chooses the control action with the longest lifetime estimate, simulates the action, counting time steps, and updates lifetime estimates depending on which action was chosen when the box was entered. The performance is understandably poor, as the discretization requires making a lose-lose trade-off between algorithmic accuracy and computation time.  That being said, other than the discretization of the state- and action-spaces, this is very similar to the algorithm we use in Algorithm \ref{VPG}.
	
	In the decades since, the emergence of reinforcement learning has lead to several other techniques, including Vanilla Policy Gradient, Actor-Critic, and Deep Q-Learning being used with varying degrees of success. The ease of use and implementation of OpenAI Gym has lead to the environment being used as a way of benchmarking the success of new algorithms like Generalized Advantage Estimation \cite{schulman}, Trust Region Policy Optimization \cite{TRPO} and Deep Deterministic Policy Gradient \cite{DDPG}. These algorithms are benchmarked on a variety of OpenAI environments in \cite{bench}. Unfortunately, few of these algorithms have been tested in real-world implementations.
	
	
	\subsubsection{Traditional Control on Real Pole}
	It is essentially impossible to exactly recreate a perfect simulation of a real world inverted pendulum in a computer. While things like friction, motor electrodynamics, and viscous damping coefficients can be included to make the simulation match reality better, various attempts to do this still neglect the (nonlinear) Coulomb friction applied to the cart, and the force on the cart due to the pendulum's action \cite{Emi,friction}. In addition, variances in manufacturing tolerances, imbalances in the setup of the cart, play in the motor, and general wear and tear on the gears all serve to add unpredictable noise that may cause simulated controls to fail, or to perform worse than the computer predicts.  Because of this significant hurdle, far fewer inverted pendulums are constructed and balanced in real life.

	A non-comprehensive list of controllers include PID with a Kalman Filter \cite{Miranda,abey}, LQR \cite{LQR1,LQR2}, State-Dependent Riccati Equation \cite{SDRE}, and power series approximation of the Hamilton-Jacobi-Bellman equation \cite{Emi}. 
	
	Proportional-integral-derivative controllers are a staple of industrial process control. By first linearizing the space state equations about the unstable equilibrium, the error $e(t)$ can be fed into the PID-controller 
	\begin{equation}
	u(t) = k_P e(t) + k_I \int_0^t e(\tau)\; d\tau + k_D \frac{d e(t)}{dt},    
	\end{equation}
	where $k_P$, $k_I$, and $k_D$ are the proportional, integral, and derivative gain constants, respectively. The performance of the controller is directly dependent on the ability to find values of these constants that work for this specific system. If any of the gain constants are too high or low, the system can become unstable, oscillate, or fail to balance the pole at all.
	
	Linear-quadratic regulator controllers also begin by linearizing the nonlinear state equations the same way, and using the solution $P$, to the algebraic Riccati equation to minimize the cost functional 
	\begin{equation}
	J(x_0, u) = \int_0^\infty x^TQx + Ru^2 \; dt.
	\end{equation}
	Then the optimal control is $u=-R^{-1}B^TPx$. Like the PID controller, the performance of LQR depends directly on the values chosen in $Q$ and $R$; this can lead to oscillations or unstable performance if they are off \cite{Tai}.
	
	In both PID and LQR, a Kalman filter is used to smooth sensor inputs, and provide an approximation of $\dot{x}$ and $\dot{\alpha}$ in the state, although a low-pass derivative filter may work better if the optimal Kalman parameter values cannot be obtained \cite{Amanda}. The downside of these controllers is that the linearized state equations differ from the true nonlinear equations increasingly as the state deviates the unstable equilibrium at the origin. While they perform well when the pole is essentially straight up, a nonlinear controller may perform better over a wider range of values, such as during a disturbance.
	
	One solution to this is to use a nonlinear formulation like the state-dependent Riccati equation. Much like LQR, SDRE starts by solving 
	\begin{multline}
	A(x)^T P(x) + P(x) A(x) \\ - P(x)B(x)R(x)^{-1}B(x)^TP(x) + Q(x)=0
	\end{multline}
	to get $P(x)\ge 0$, which is used for the control $u = -R(x)^{-1}B(x)^TP(x)x$. The main difference between LQR and SDRE is that the design matrices $Q(x)$ and $R(x)$ and the plant matrices $A(x)$ and $B(x)$ are state-dependent instead of constant. This allows the same controller to be used to swing-up and balance the pendulum, instead of switching controllers, as the linearized models would require. The major downside of this technique is that it requires solving the algebraic Riccati equation at every timestep, which can be computationally expensive and limit the sampling rate, but results in a controller that is more stable and robust.

	\subsubsection{Reinforcement Learning on Real Pole}
	Due to the additional challenge presented by balancing a real pole, there have even been several attempts at using various reinforcement learning schemes to control a real world inverted pendulum. They all suffer from the same issue: most reinforcement algorithms require a lot of data to perform adequately, and getting data from a real pendulum is a slow and expensive process. Admittedly, some algorithms are better than others, requiring only a few minutes of ``online'' time, interacting with the pendulum, but the total time training can be an order of magnitude higher.
	
	One technique that that performs well is called Neural-Fitted Q Iteration (NFQ) \cite{neural}, which does not require a model and is relatively data efficient. This technique is a modification of a tree-based regression method using neural networks with an enhanced weight update method that can learn directly from real-world interactions in the form (state, action, successor state). This is a type of Markov Decision Process described by a set of states $S$, actions $A$, and a process $p(s_t, a_t, s_{t+1})$ for describing transition behaviour. The goal is to find an optimal policy $\pi^*: S \rightarrow A$ that minimizes expected costs (or maximizes expected rewards) for each state.
	
	Classical Q-learning uses the update rule
	\begin{multline}
	Q_{k+1}(s_t,a_t) = (1-\alpha_r)Q(s_t,a_t)\\+ \alpha_r \left(c(s_t,a_t) + \gamma \min_a Q_k(s_{t+1},a)\right),
	\end{multline}
	where $\alpha_r$ is the learning rate, $c(s,a)$ is the cost of taking action $a$ in state $s$, and $\gamma$ is the discount factor. When $S$ is finite, a table of Q-values can be made and updated using this equation. When $S$ is continuous, an error function like $\left(c(s_t,a_t) + \gamma \min_a Q_k(s_{t+1},a)\right)^2$ can be introduced and minimized using backpropagation and gradient descent. Again, this is essentially just a reformulation of the algorithm we use in Algorithm \ref{VPG}.
	
	However, it has the specific downside of requiring many iterations (potentially tens of thousands) until a near-optimal policy is found, which is unrealistic to obtain for many agents. Simulation, and larger learning rates can help alleviate this issue. Alternatively, Neural-Fitted Q Iteration uses offline learning, enabling more advanced batch supervised learning algorithms like Rprop to use the entire set of transition experiences when updating the neural network parameters. This drastically reduces the amount of data required, proportionally reducing training time on a real agent or in simulation. As a bonus, Rprop is insensitive to the learning parameters, meaning time-consuming hyperparameter optimization is not necessary to achieve quick learning.
	
	Alternative data-efficient model-based strategies like Probabilistic Inference for Learning Control (PILCO) \cite{PILCO} are also able to learn effective policies with very little data by incorporating model uncertainty with a probabilistic dynamics model for long-term planning. Essentially, this algorithm creates a probabilistic model using fewer data points, and refines the model as more data is collected. Ultimately, this results in a drastic reduction in required training to achieve satisfactory results.
	
	Given a sample, PILCO uses a squared exponential kernel to define a Gaussian posterior predictive distribution. By mapping uncertain test inputs through the Gaussian process dynamics model, it produces a Gaussian approximation of the desired posterior distribution, which can be used to analytically compute the gradients to minimize the expected return, through standard algorithms like gradient descent, conjugate gradient, or limited-memory Broyde-Fletcher-Goldfarb-Shanno algorithm.

	\subsection{Preliminaries}
	Most cartpole implementations are either simulations using simplified models, or model-free implementations in the real world. This research is unique and novel in that we can balance real-world pendulums using neural networks trained entirely on a computer using a mechanistic model.
	
	\subsubsection*{Environment}
	The environment used was modified from OpenAI\footnote{\url{https://openai.com/}} Cartpole, which has the same setup as in Figure~\ref{SIP}, except the motor can only apply a force of $-1$ or $+1$, moving the cart left or right, with no consideration for how fast or how far to move. Additionally, the simplistic physical model was updated using equations for horizontal and angular acceleration from \cite{Emi} that were created to be as physically accurate as possible, by incorporating friction, motor electrodynamics, and the viscous damping coefficient as seen at the motor pinion and pendulum axis. Finally, the update scheme was updated to a modified version of semi-implicit Euler's method, which almost conserves energy, and prevents the system from becoming unstable over long time intervals.
	
	The equations of motion are given by:
	
	\begin{equation}
	\begin{split}
	\ddot{x}(t) = &-\dfrac{3r_{mp}^2 B_p \cos(\alpha(t))\dot{\alpha}(t)}{\ell_p D(\alpha)} \\ & -\dfrac{4M_p \ell_p r_{mp}^2 \sin(\alpha(t))\dot{\alpha}(t)^2}{D(\alpha)} \\ & -\dfrac{4(R_m r_{mp}^2 B_{eq}+K_g^2 K_t K_m)\dot{x}(t)}{R_m D(\alpha)} \\ & + \dfrac{3M_p r_{mp}^2 g \cos(\alpha(t))\sin(\alpha(t))}{D(\alpha)}  \\ & + \dfrac{4r_{mp} K_g K_t V_m}{R_m D(\alpha)},\\
	\ddot{\alpha}(t) = &-\dfrac{3(Mr_{mp}^2 + M_p r_{mp}^2 +J_m K_g^2)B_p\dot{\alpha}(t)}{M_p \ell_p^2 D(\alpha)}  \\ & -\dfrac{3M_p r_{mp}^2 \cos(\alpha(t))\sin(\alpha(t))\dot{\alpha}(t)^2}{D(\alpha)} \\ &  -\dfrac{3(R_m r_{mp}^2 B_{eq}+K_g^2 K_t K_m)\cos(\alpha(t))\dot{x}}{R_m \ell_p D(\alpha)}  \\ & +\dfrac{3(Mr_{mp}^2 + M_p r_{mp}^2+ J_m K_g^2)g \sin(\alpha(t))}{\ell_p D(\alpha)} \\ &+\dfrac{3r_{mp} K_g K_t \cos(\alpha(t))V_m}{R_m\ell_p D(\alpha)},
	\end{split} \label{EOM}
	\end{equation}
	where $D(\alpha) = 4Mr_{mp}^2 + M_p r_{mp}^2 + 4 J_m K_g^2 + 3M_p r_{mp}^2 \sin^2(\alpha(t))$, and the position is updated via semi-implicit Euler's method:
	\begin{equation}
	\begin{split}
	\dot{x}_{t+1} & = \dot{x}_t + h \ddot{x}_t,\\
	\dot{\alpha}_{t+1} &= \dot{\alpha}_t + h\ddot{\alpha}_t,\\
	x_{t+1} & = x_t + h\dot{x}_t,\\
	\alpha_{t+1} & = \alpha_t + h\dot{\alpha}_t,
	\end{split} \label{Euler}
	\end{equation}
	with time-step size $h = \frac{1}{50}$.
	
	See Section 2.2 of \cite{Emi} for details.
	
	\section{\uppercase{METHODOLOGY}}
	
	\subsection{Rewards}
	The goal of reinforcement learning is to make an agent choose the optimal policy to maximize the rewards it will receive. In this case, our agent is the inverted pendulum, the policy is a one-hidden-layer artificial neural network that decides how much voltage to apply to the motor, and the rewards correspond to how long the pole stays balanced.
	
	We define the reward function as an initially empty vector $r = [\;]$. For every time-step of length $h$, we calculate the new state $s_{t+1}$ given the current action $a_t$, state $s_t$, and Equations~\eqref{EOM} and \eqref{Euler}. If the agent survives (the pole remains between $\pm12^{\circ}$ and stays within the bounds of the track), we set $r_{t+1} = 1$.  Given a trajectory of states and actions $\tau = \left(s_0, a_0, \dots, s_{T+1}\right)$, our goal is to choose the policy $\pi_\theta$ that allows the pole to remain balanced by maximizing
	\begin{equation}
	R(\tau) = \sum_{t=0}^T r_t.
	\end{equation}
	
	This sum diverges over infinite timescales. A common technique to prevent this (and to emphasize getting rewards now rather than later in the future), involves the idea of a discounted reward. That is, we define $\gamma \in(0,1)$ so that our cumulative discounted reward at time step $t$ is:
	
	\begin{equation}
	R_t = \sum_{k=0}^\infty \gamma^k r_{t+k}.
	\end{equation}
	In addition to guaranteeing the sum converges, discounting has the additional benefit of prioritizing rewards now over rewards later. The value of $\gamma$ can be tuned to provide the optimal combination of current and future rewards. Small values of $\gamma$ prioritize actions resulting in rewards now, while larger values prioritize rewards over a longer time frame. If we normalize $R$ by subtracting the mean and dividing by the standard deviation, we can simultaneously encourage and discourage roughly half of the possible actions, as seen in \cite{Karpathy,schulman}.
	
	\subsection{Loss}
	Most machine learning algorithms seek to minimize the loss function through gradient descent or one of its variants. Since our goal is to maximize the likelihood of actions that result in large rewards, we can do this by \textit{minimizing} the negative likelihood (or negative log-likelihood, since the natural logarithm is monotonically increasing). We define our loss as
	\begin{equation}
	L = -R\cdot \tilde{a} + \epsilon H,
	\end{equation}
	where $R$ is the vector of normalized discounted rewards, $\tilde{a}$ is the the vector of log-likelihoods of the actions taken (as determined by the mean and variance), and $\epsilon$ is a regularization parameter augmenting the weight of $H$ -- the amount of Shannon entropy. This optional parameter allows the entropy $H = \ln(\sigma\sqrt{2\pi e})$ to prevent the standard deviation from getting too small, preventing exploration of moves in the action-space.

	\subsection{Policy Gradient}
	Now that we've defined the agent, state, rewards, and loss function, let's put it all together. Vanilla Policy Gradient is one of the simplest methods in reinforcement learning \cite{OpenAI}, meaning there is plenty of future work to be done, incorporating a variety of data-efficient reinforcement learning algorithms from Section~\ref{lit}.
	
	Let $\pi_\theta$ be a policy with parameters $\theta$ and define $J(\pi_\theta) = \underset{\tau\sim\pi_\theta}{E} \left[R(\tau)\right]$ be the expected return of the policy. Given a trajectory of states and actions $\tau = \left(s_0, a_0, \dots, s_{T+1}\right)$, we define the probability of $\tau$ given actions sampled from $\pi_\theta$ to be
	\begin{equation}
	P\left(\tau\vert\theta\right) = \rho_0 \prod_{t=0}^T P(s_{t+1}\vert s_t,a_t)\pi_\theta(a_t,s_t).
	\end{equation}
	
	Then the gradient of $J(\pi_\theta)$ is
	
	\begin{equation}
	\begin{split}
	\nabla_\theta J(\pi_\theta) & = \nabla_\theta \underset{\tau\sim\pi_\theta}{E} \left[R(\tau)\right]\\
	& = \nabla_\theta \int_\tau P(\tau\vert\theta)R(\tau)\\
	& = \int_\tau \nabla_\theta P(\tau\vert\theta)R(\tau)\\
	& = \int_\tau P(\tau\vert\theta) \nabla_\theta\log P(\tau\vert\theta)R(\tau)\\
	& = \underset{\tau\sim\pi_\theta}{E} \left[\nabla_\theta \log P(\tau\vert\theta)R(\tau)\right].
	\end{split}
	\end{equation}
	
	At this point, we can choose any arbitrary advantage function, $A^{\pi_\theta}$, relative to the current policy, depending on the current action and state, giving us
	\begin{multline}
	\nabla_\theta J(\pi_\theta) = \\\underset{\tau\sim\pi_\theta}{E} \left[\sum_{t=0}^T \nabla_\theta \log \pi_\theta (a_t\vert s_t) A^{\pi_\theta}(s_t,a_t)\right].
	\end{multline}
	
	In general, we should choose $A^{\pi_\theta}(s_t,a_t)$ so that the gradient term $\nabla_\theta \log \pi_\theta (a_t\vert s_t) A^{\pi_\theta}(s_t,a_t)$ points in the direction of increased $\pi_\theta (a_t\vert s_t)$ if and only if $A^{\pi_\theta}(s_t,a_t)>0$. Therefore, we choose $A^{\pi\theta}$ to be our normalized discounted reward $R_t$ from above (although other choices are possible and may be beneficial for certain applications \cite{schulman}).
	
	At this stage, we can simply proceed through general gradient descent using whatever optimizer we choose.
	
	\subsection{Training}
\begin{figure}[!hbt]
		
		\begin{center}
			\mbox{
				\includegraphics[width=\columnwidth]{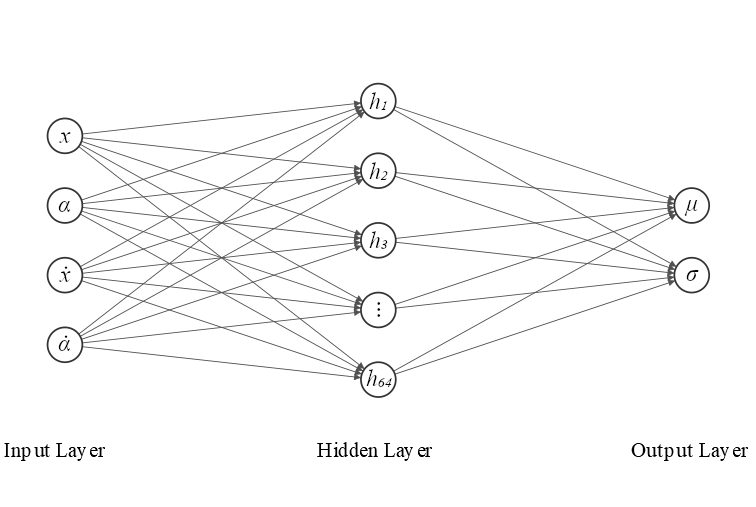}
			}
		\end{center}
		
		\caption{$\pi_\theta$ is defined by an artificial neural network.}
		\label{Network}	
	\end{figure}

	We define $\pi_\theta$ as a one-hidden-layer artificial neural network as depicted in Figure~\ref{Network}. The input layer consists of four neurons, corresponding to the state vector $[x,\alpha,\dot{x},\dot{\alpha}]$. This feeds forward to the 64-neuron hidden layer, with \texttt{ReLU} activation. 64 neurons was chosen somewhat arbitrarily, based on results from the literature and early simulations conducted without friction. While 32 neurons is enough for the agent to learn to balance itself, more neurons (and more layers) may result in faster training speed. The top layer consists of two neurons defining the normal probability distribution $N(\mu, \sigma)$ the policy samples from. The first neuron uses linear activation and corresponds to the mean of the distribution, while the second neuron uses softplus ($\ln(1+e^x)$) activation (guaranteeing it will be positive) and corresponds to the standard deviation of the normal distribution.

	As seen in Algorithm~\ref{VPG}, the model is fed the current state $s_t$ and outputs a probability distribution of actions to be taken. The policy samples an action $a_t$ from the distribution, clips it to $[-10,10]$ due to the physical limitations of the motor, and uses the Equations of Motion~\eqref{EOM} and \eqref{Euler} to step forward in time, generating the new state $s_{t+1}$. If the pole remains balanced (angle between $\pm12^\circ$ and the cart is still on the track), then $r_{t+1}=1$ is appended to the rewards vector and training continues. If the model fails to meet these criteria, training stops, the rewards are discounted and normalized, and the weights are updated through backpropagation. We found the Adam optimizer, with a learning rate of $0.01$ tended to perform the best.
	
	Training is considered finished when the agent performs $500$ time steps (balancing for 10s) without crashing. Depending on the hyperparameters chosen for the learning rate and $\gamma$, this can take anywhere from $54$ individual trials (taking {\raise.17ex\hbox{$\scriptstyle\sim$}}1 minute on Google Colab's Tesla K80 GPUs) to well over $2000$. At this point, optional additional training with a smaller learning rate can help in making the model more robust.
	
	\begin{algorithm}[hbt]
		initialize state $s_t$ so $|x|<0.4$ and $|\theta|<12^\circ$\;
		initialize $r=0$\;
		\While{$r<500$ time steps}{
			choose action by sampling $a_t\sim\pi(\cdot\vert s_t)$\;
			clip $a_t$ to $[-10,+10]$\;
			calculate $s_{t+1}$ using Equations of Motion\;
			\eIf{$|x|>0.4$ OR $|\alpha|>12^\circ$}{
				Crashed\;
				$r=0$\;
				go to Training\;
			}{
				$r:=r+1$\;
				go to start of while loop\;
			}
			Training\;
			compute log-likelihoods $\tilde{a}$\;
			normalize discounted rewards $R$\;
			compute loss $=-R\cdot\tilde{a}$\;
			update parameters using with gradient descent $\left(\theta := \theta - \alpha \nabla_\theta J(\pi_\theta)\big\vert_\theta\right)$ or Adam optimizer\;
		}
		Training is finished when $r=500$
		\caption{Training with Vanilla Policy Gradient}
		\label{VPG}
	\end{algorithm}
	\vspace{0.3in}
	
	\section{\uppercase{EXPECTED OUTCOME}}
	
	Once a policy has been learned, the neural network can be downloaded as a series of matrices $W_1$, $b_1$, $W_2$, $b_2$ corresponding to the weights and biases of each layer of the neural network. These can be loaded into MATLAB, and the network recreated in a single block in Simulink. While later versions of MATLAB have neural network support, the version we are using does not, meaning it must be recreated from scratch.
	
	\subsection*{Real World Pendulum}
	At this point, the model can be built with Quanser QUARC, loaded to the controller, and tested on a real-world pendulum. Preliminary results in the lab show this to be working extremely well. Models that succeeded in balancing in the simulation for the full 10 seconds are robust enough to balance for much longer than that in simulation, and can replicate this success in the real-world. One trial lasted well over 5 minutes, without large or jerky movements, and was able to correct course after being gently tapped in either direction without falling.
	
	We trained several dozen simulated models to success, each of which behaved slightly differently, given a random initial state (tending to drift in one direction, large oscillations in the middle of the track, jerky movements, etc.). When deployed on the real pendulum, they tended to replicate those movements.

	\section{\uppercase{STAGE OF THE RESEARCH}}
	Using the Policy Gradient algorithm, we have successfully trained a simulation of a single inverted pendulum to balance on its own. Across a variety of learning rates and discount factors tested, the pole balanced for 10 seconds 91\% of the time, taking an average of 807 trials in order to get there. Due to the large variance of the policy gradient algorithm, hyperparatmer optimization was not able to find a single set of parameters that performed optimally. However, this same process found several hyperparamter configurations that performed poorly and were eliminated from consideration. The pole averaged 355 trials to balance among those that remained. The best result balanced the pendulum in only 54 trials.
	
	\subsection{Current Results}
\begin{figure*}[htb]
		\centering
		\includegraphics[width=0.25\textwidth]{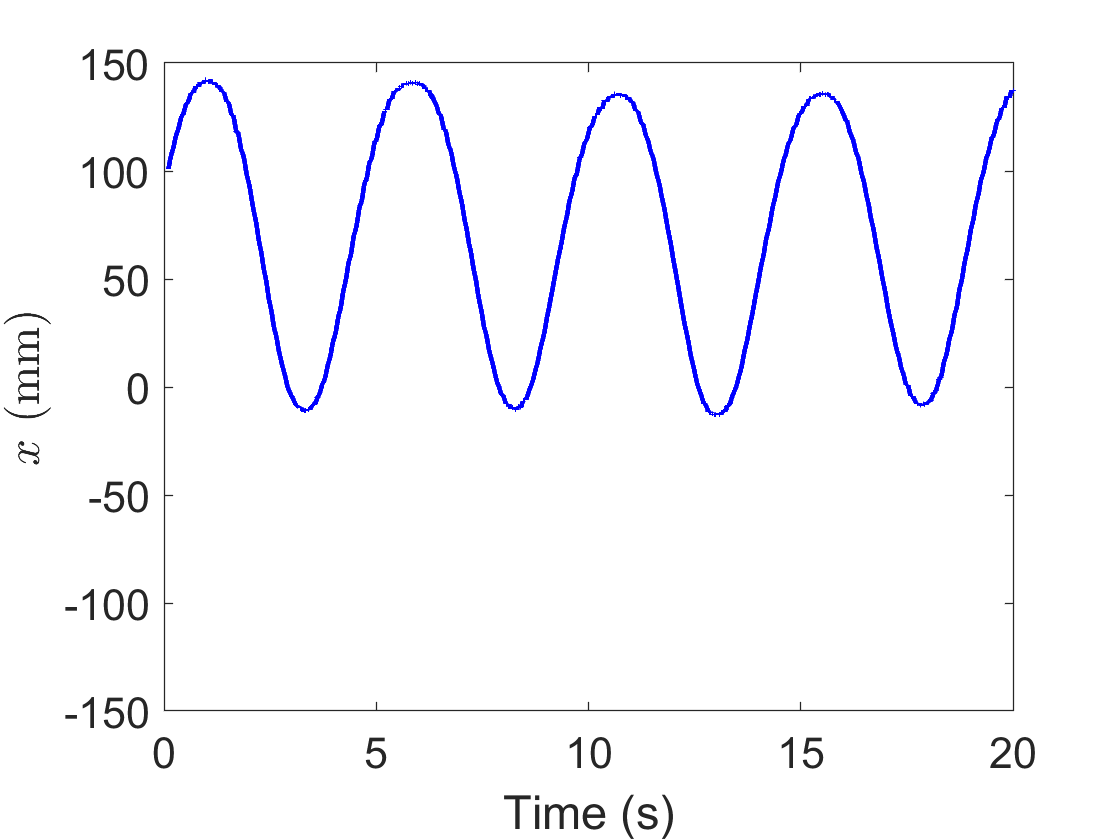}\includegraphics[width=0.25\textwidth]{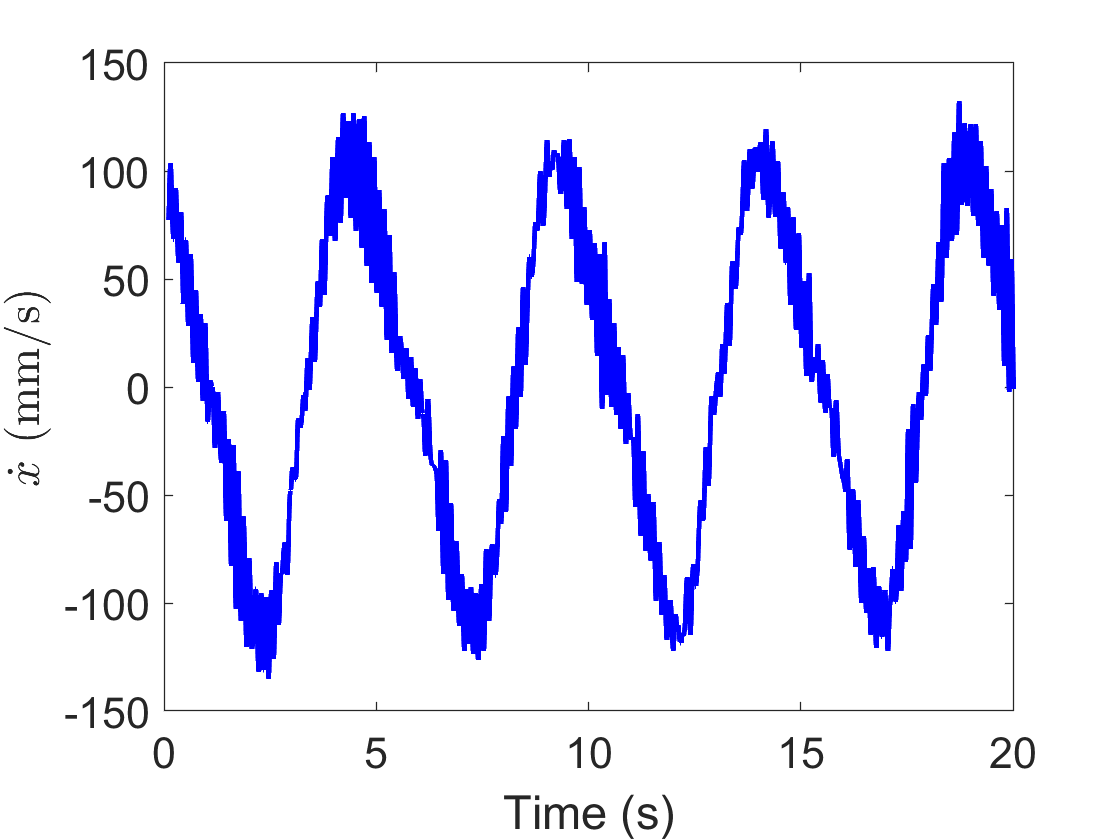}\includegraphics[width=0.25\textwidth]{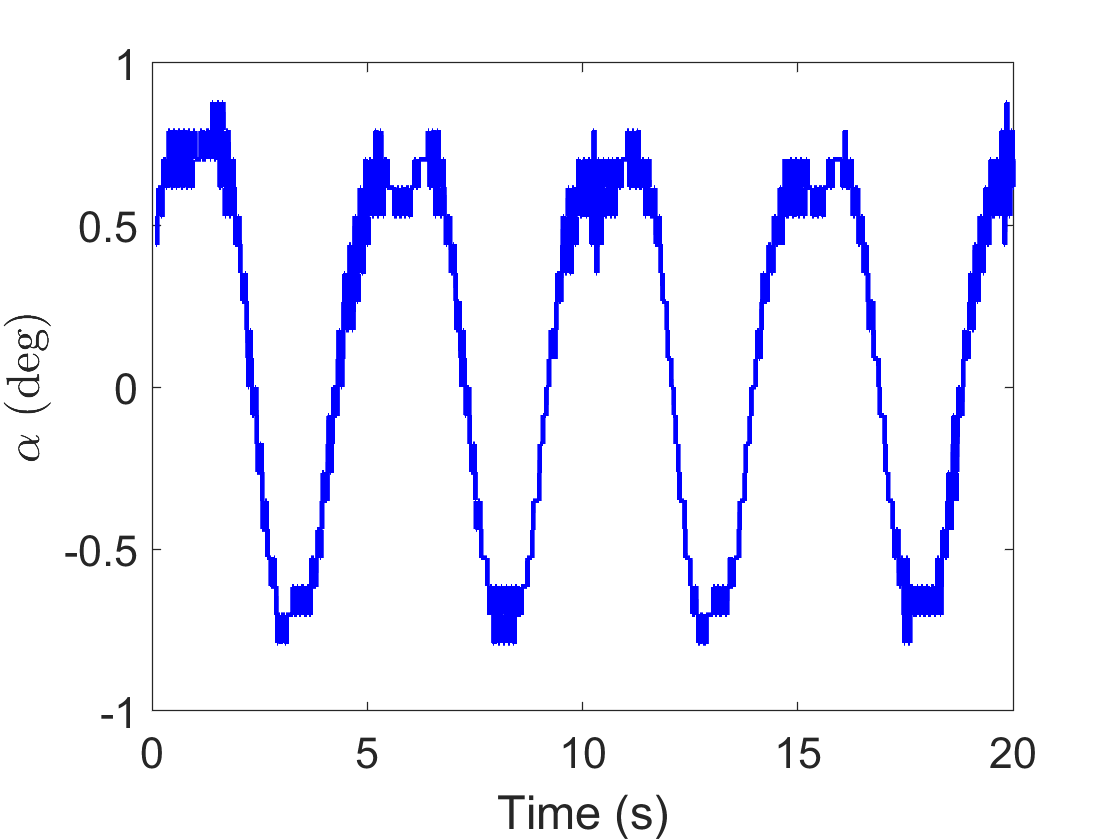}\includegraphics[width=0.25\textwidth]{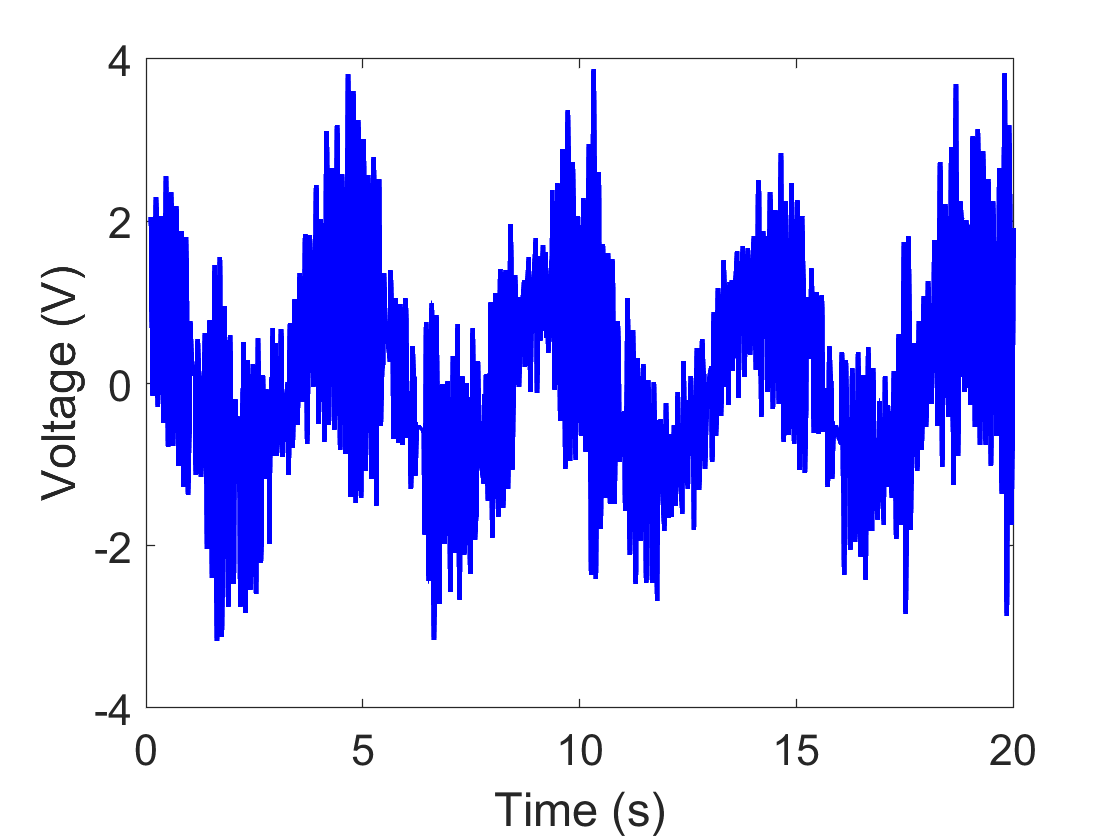}
		\includegraphics[width=0.25\textwidth]{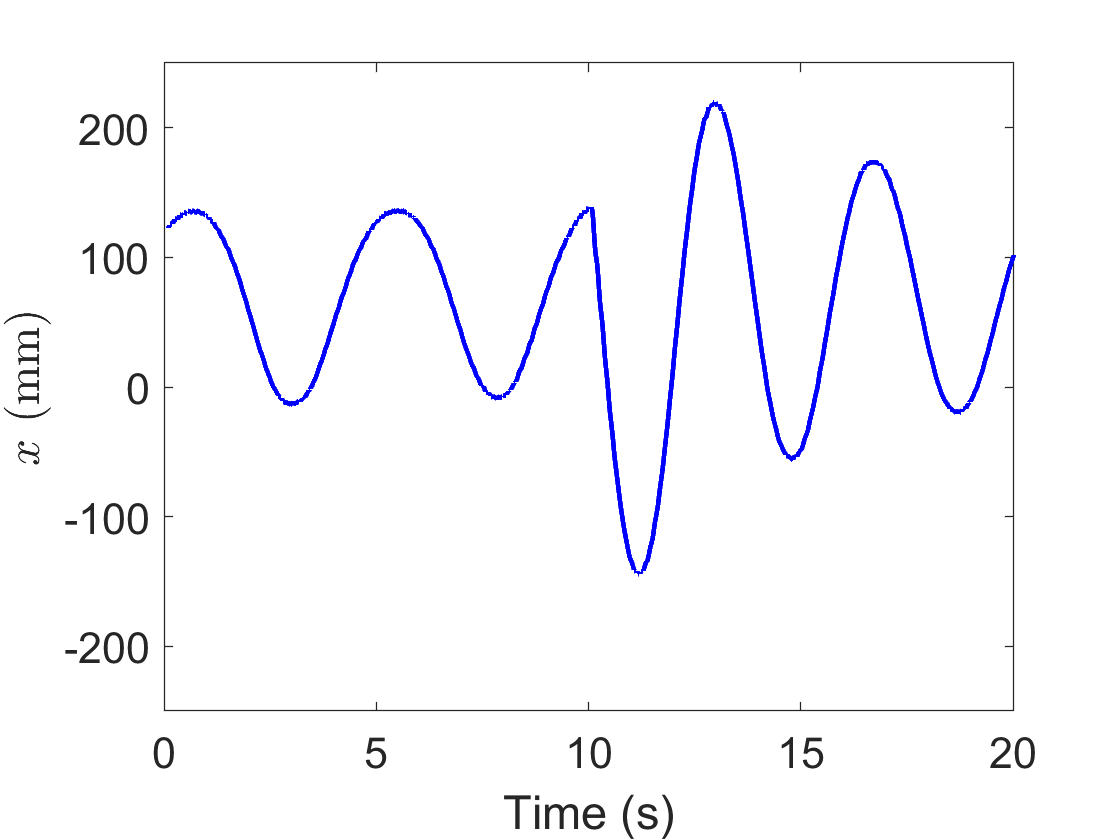}\includegraphics[width=0.25\textwidth]{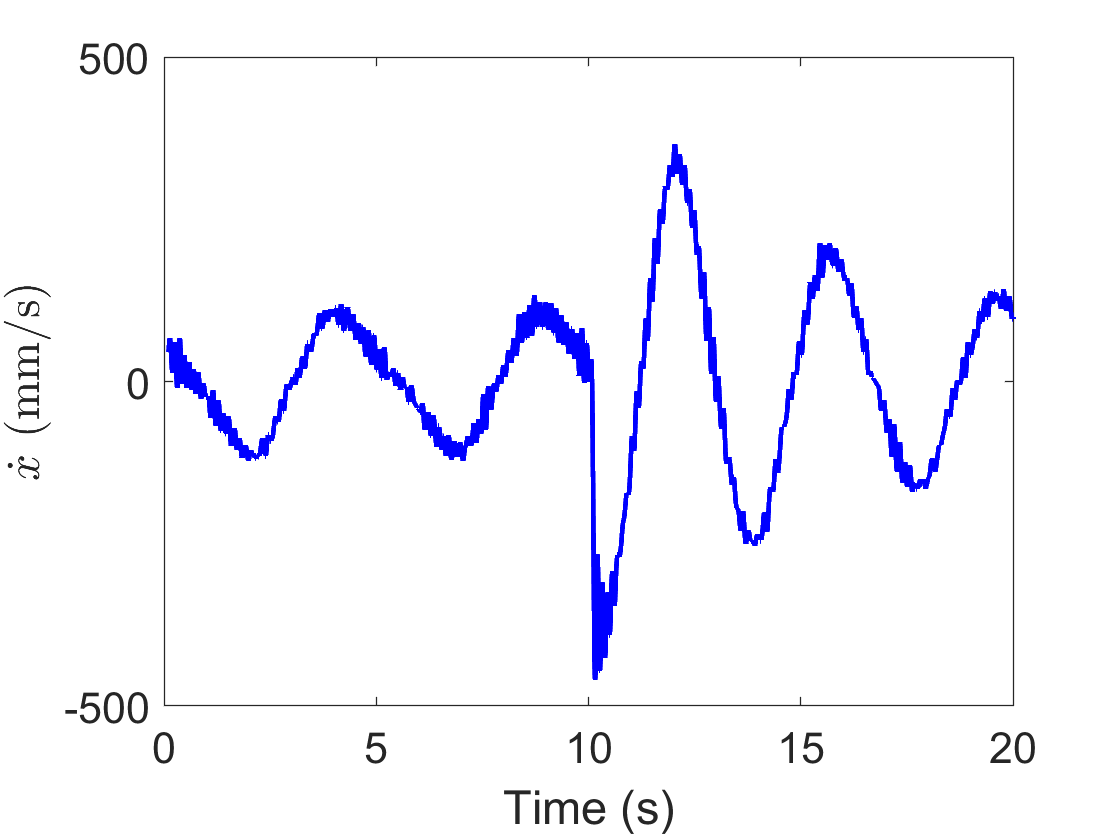}\includegraphics[width=0.25\textwidth]{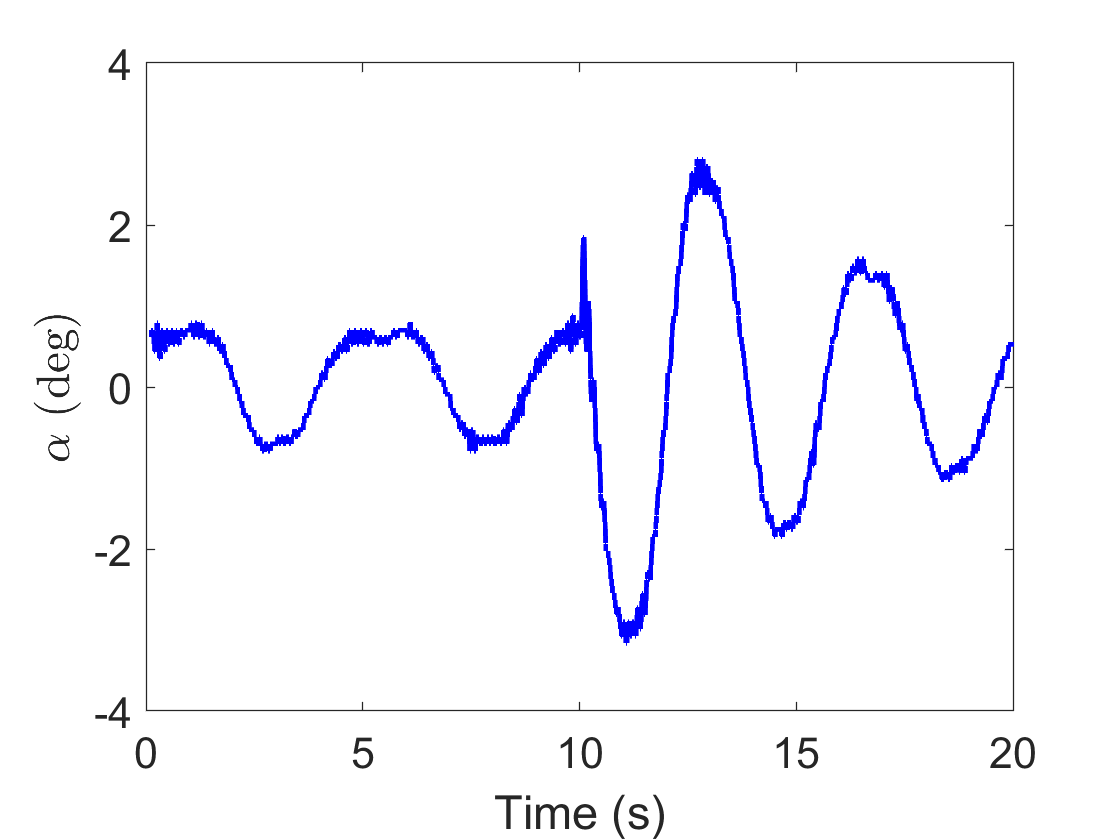}\includegraphics[width=0.25\textwidth]{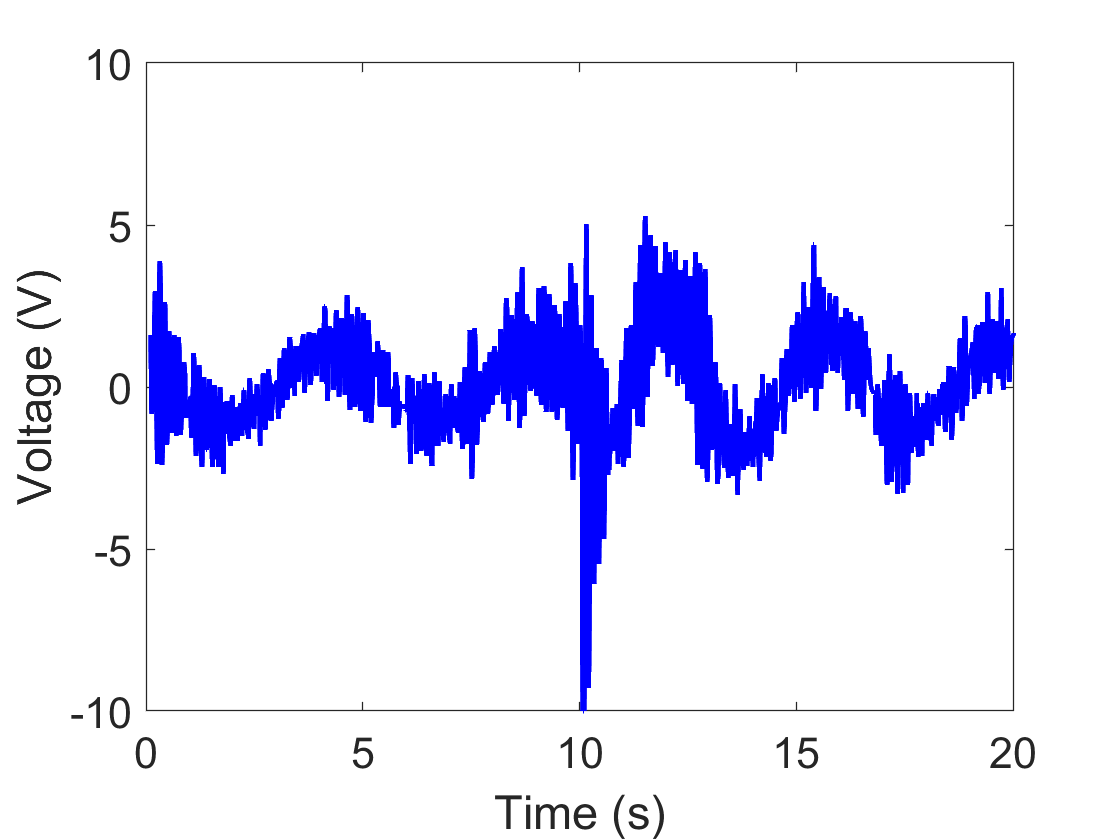}
		\caption{Position (mm), velocity (mm/s), angle (deg), and voltage (V) of the agent over a 20 second timeframe unperturbed (top) and with a disturbance (a light tap on the pole) 10 seconds in (bottom).}
		\label{graphs}
	\end{figure*}

	As seen in Figure~\ref{graphs}, a virtually-trained policy is able to balance a real inverted pendulum, even when the pole is disturbed (i.e. by tapping it). While the cart naturally oscillates left and right (first column), the pole never exceeds 1 degree from vertical (third column) and the motor never reaches saturation (fourth column), except when disturbed (bottom row). That being said, the cart's motion is relatively smooth compared to implementations like \cite{neural}\footnote{\url{https://www.youtube.com/watch?v=Lt-KLtkDlh8}} that used reinforcement learning directly on a real pendulum, that also take longer to train. 
	
	As we have shown, it is possible to harness GPUs to quickly train a simulation of an inverted pendulum to balance itself that performs comparably in the real world. This could be used for other data-sparse, time-intensive reinforcement learning applications that can be accurately modeled or simulated. The requirement of accurate simulation means that model-free or stochastic processes cannot be used. However, even simplified models may be sufficient to train an agent. A simplified simulation of the physics of an inverted pendulum without friction was still able to quickly learn a policy that can balance in the real world. Additionally, the real pendulum was still able to balance for a short period of time when the length of the pole and mass of the cart were changed.

	\subsection{Future Work}
	Because of the large variance resulting from the random initialization, the agent sometimes takes more than an order of magnitude longer to train during unlucky initializations. Optimizations in the training process, such as using physically-based rewards, specific to this agent and environment may reduce this variance. Additionally, alternative reinforcement learning algorithms like Generalized Advantage Estimation \cite{schulman}, Trust Region Policy Optimization \cite{TRPO} and Deep Deterministic Policy Gradient \cite{DDPG} have been benchmarked on a variety of virtual OpenAI environments in \cite{bench}, but few of them have been tested in real-world implementations. Data-sparse algorithms like NFQ or PILCO may help to reduce variance and regularize the training process, with the added benefit of reducing training time on top of the efficiency already provided by training on a computer.
	
	Moving forward, it would be interesting to see how this algorithm works on alternative applications like swinging up the pendulum, or on alternative agents like a double inverted pendulum. Different reinforcement learning algorithms may perform differently for different learning tasks \cite{bench}.
	
	The idea of training models in simulation for real-world implementation is already seeing use in research labs \cite{Rubik}. Researchers at OpenAI are using two interconnected long short-term memory neural networks defining a value function and agent policy to be trained in simulation and implemented in the real world to solve a Rubik's cube using a robotic hand. They are using Proximal Policy Optimization to train the control policies and use automatic domain randomization to train on a large distribution
	over randomized environments. Despite not being able to model accurate physics in the simulations, domain randomization allows them to successfully solve a cube 60\% of the time.
	
	Learning from their success, domain randomization could potentially allow us to balance poles of arbitrary length and weight (within the constraints the motor could handle) without requiring specific knowledge of the various physical attributes of the pole we currently use.

	\section*{\uppercase{ACKNOWLEDGEMENTS}}
	I would like to thank my advisor, Dr. Hien Tran, for his constant advice, insight, and support and I would also like to thank my cat. Funding support for this research was provided by the Center for Research in Scientific Computation.
	\nocite{quanser}
	\bibliographystyle{apalike}
	{\small
		\bibliography{Example}}

	\section*{\uppercase{APPENDIX}}
	Table~\ref{tab:parameters} contains the model parameter values used for the Equations of Motion in simulation. Some of these values found in the technical specifications of the Quanser User Manual (Quanser, 5.0) were found to be incorrect, and were determined experimentally through parameter identification in \cite{Emi}.
	
	\begin{table*}[tb]
		\caption{Model parameter values used in Equation~\ref{EOM}.}\label{tab:parameters} \centering
		\begin{tabular}{|c|c|c|}
			\hline
			Parameter & Description & Value \\
			\hline
			$B_p$ & Viscous Damping Coefficient, as seen at the Pendulum Axis & $\unit{0.0024} \newton\usk\metre\usk\second\per\radian$  \\
			\hline
			$B_{eq}$ & Equivalent Viscous Damping Coefficient as seen at the Motor Pinion & $\unit{5.4}\newton\usk\metre\usk\second\per\radian$ \\
			\hline
			$g$ & Gravitational Constant & $\unit{9.8} \metre\per\second\squared$ \\
			\hline
			$I_p$ & Pendulum Moment of Inertia, about its Center of Gravity & $\unit{8.539\times 10^{-3}} \kilogram\usk\metre\squared$ \\
			\hline
			$J_p$ & Pendulum's Moment of Inertia at its Hinge & $\unit{3.344\times 10^{-2}} \kilogram\usk\metre\squared$\\
			\hline
			$J_m$ & Rotor Moment of Inertia & $\unit{3.90\times 10^{-7}}\kilogram\usk\metre\squared$ \\
			\hline
			$K_g$ & Planetary Gearbox Ratio & $3.71$ \\
			\hline
			$K_t$ & Motor Torque Constant & $\unit{0.00767} \newton\usk\metre\per\ampere$ \\
			\hline
			$K_m$ & Back Electromotive Force (EMF) Constant & $\unit{0.00767} \volt\usk\second\per\radian$ \\
			\hline
			$\ell_p$ & Pendulum Length from Pivot to Center Of Gravity & $\unit{0.3302} \metre$ \\
			\hline
			$M$ & Cart Mass & $\unit{0.94} \kilogram$ \\
			\hline
			$M_p$ & Pendulum Mass & $\unit{0.230} \kilogram$ \\
			\hline
			$R_m$ & Motor Armature Resistance & $\unit{2.6} \ohm$ \\
			\hline
			$r_{mp}$ & Motor Pinion Radius & $\unit{6.35\times10^{-3}} \metre$ \\
			\hline
		\end{tabular}
	\end{table*}
\end{document}